\documentclass[11pt,a4paper]{article}
\usepackage[utf8]{inputenc}
\usepackage[T1]{fontenc}
\usepackage{newtxtext,newtxmath}
\usepackage[margin=25mm]{geometry}
\usepackage{amsmath}
\usepackage{graphicx}
\usepackage{booktabs}
\usepackage{caption}
\usepackage{textcomp}
\usepackage[affil-it]{authblk}
\usepackage[colorlinks=true,linkcolor=black,citecolor=black,urlcolor=blue!50!black]{hyperref}
\graphicspath{{figures/}}

\title{\LARGE When Magnitude Suffices: Reconstructing Time-Domain-Gated Microwave Signals Without Phase on a Thoracic Phantom}
\author[a]{Maciej Cicho\'n\thanks{Corresponding author: maciej.cichon@fis.uni.lodz.pl}}
\author[a]{Patrycja Go\l{}a\'s}
\author[a]{Maksymilian Bartosik}
\author[a]{Angelika Walczak}
\author[b]{Agata Bielecka-D\k{a}browa}
\author[a]{Maciej \'Slot}
\affil[a]{Department of Solid State Physics, Faculty of Physics and Applied Informatics, University of \L{}\'od\'z, Pomorska 149/153, 90-236 \L{}\'od\'z, Poland}
\affil[b]{Polish Mother's Memorial Hospital Research Institute, Rzgowska 281/289, 93-338 \L{}\'od\'z, Poland}
\date{}

\begin{document}
\maketitle

\begin{abstract}
\noindent Time-domain gating separates the arrivals that form a microwave S-parameter measurement. It requires the measurement of complex values, so this application almost always uses a vector network analyzer (VNA). We show that, on one calibrated measurement configuration, the gated magnitude the VNA would give can be reconstructed from a magnitude-only sweep of a spectrum analyzer with a tracking generator (SA). We select the gate center for separability of the measured states, correct each instrument once by a per-bin subtraction of a reference state, and train a fully connected network on sweeps from both instruments. The test object is a torso phantom with lung-shaped compartments whose filling takes eleven dielectric states, each held out in turn. On $S_{11}$ the reconstruction from SA input reaches a shape error of 0.08~dB$^2$ with a state ordering of 0.81. This matches the reconstruction from VNA input. On $S_{21}$ it reaches 0.82~dB$^2$ and keeps the ordered groups of the states. An explicit phase prediction performs much worse, because gating needs the phase to about 0.2~rad RMS. Across eleven measurement configurations the dependence of phase on magnitude is strong for reflection and weak for transmission. This explains the difference between $S_{11}$ and $S_{21}$. Where the magnitude carries enough of the phase, a scalar instrument gives the gated response without any phase measurement after installation. On the open UM-BMID breast data set, with whole phantoms held out, the shape error of the network is 30 percent lower than that of a per-bin linear map.

\medskip
\noindent\textbf{Keywords:} magnitude-only measurement; scalar network analysis; time-domain gating; spectrum analyzer; microwave sensing; thoracic phantom; neural network regression
\end{abstract}

\section{Introduction}

Microwave measurements in the low-GHz band respond to the dielectric state of tissue. Soft tissues differ enough in permittivity \cite{r1} that fluid collecting in the thorax changes the S-parameters of an antenna pair held at the body surface. The radiation is non-ionizing, and the measurement needs no shielded room and no specialized imaging staff. Systems built on this principle range from multi-antenna torso scanners \cite{r2,r3} to deep-tissue tumor detection \cite{r4,r5,r6} and to the simplified body-mounted transmission geometry that has reached clinical tests for pulmonary congestion \cite{r7,r8,r9}. Microwave assessment of lung water was demonstrated as early as 1979 \cite{r10}, wideband systems for non-invasive monitoring of pulmonary edema followed \cite{r11}, and the field is surveyed in reviews of electromagnetic torso scanning \cite{r12} and of microwave medical sensing and imaging \cite{r13}.

With the antennas outside the body, the receiver sees several arrivals at once: the antenna's own mismatch, returns from the cabling, energy that travels around the body, and the part that has passed through the region we want to interrogate. They arrive at different times, but at each frequency the instrument reports their sum, so in the frequency domain they cannot be told apart. Time-domain gating is the standard way to separate them: the complex S-parameter is transformed into the time domain; a window keeps the interval that holds the arrival of interest and the rest is discarded. Gating is therefore what makes the measurement interpretable, and it is also what sets the instrument. Time-domain gating requires the complex S-parameter spectrum, including both magnitude and phase. In conventional measurements, this information is provided by a vector network analyzer (VNA). The requirement for a VNA therefore arises from the signal-processing procedure rather than from the underlying physical interaction itself. This raises a practical question: how firm is this requirement?

The answer depends on how much of the phase the magnitude already carries. The two are measured separately, but they are not free of each other. The impulse response of one of these measurements is short. A few arrivals, each with a delay and an amplitude. Those few numbers generate the whole spectrum at every frequency. In the limiting case of a minimum-phase system, the phase can be derived from the magnitude response through the Hilbert-transform relation \cite{r14}. A measurement that contains a delayed arrival is not minimum phase, so the magnitude does not fix the phase completely. How close it comes depends on the antennas, the geometry, and which arrivals compete. It is a property of one measurement configuration, not a constant, and it can be measured before any reconstruction is attempted. Where the magnitude comes close enough, the gated output can be recovered from a magnitude-only sweep, and the phase never has to be measured again after installation.

What is gained by not measuring phase is more than the price difference between the two instruments. Phase is the demanding half of a network measurement. It needs a coherent reference channel, a vector calibration to fix a reference plane, cables whose electrical length does not change when they are moved or warmed, and a sweep slow enough for the source to settle at every point. None of this is needed for magnitude. Each of them is also a way for an instrument in the field to drift out of specification without the operator seeing anything wrong in the data. A spectrum analyzer (SA) equipped with a tracking generator can provide a frequency-dependent magnitude response without directly measuring phase and without requiring a coherent reference channel. Due to this fact it does not need through-calibration and has no reference plane to lose. It reports only $|S(f)|$, so a raw SA sweep cannot be gated. In this work we show how to remove this limitation in the context of lung fluid detection in lossy phantom measurements.

Scalar network analyzers and spectrum analyzers with a tracking generator report $|S(f)|$ without a reference channel. The classical method from such data to a complex result is the minimum-phase relation \cite{r14}. In this relation the phase follows from the magnitude through the Hilbert transform when the system has no excess delay. A measurement that contains delayed arrivals is not minimum phase. Section~6 reports what the relation gives in that case.

Analytical methods recover error-corrected two-port S-parameters from uncalibrated measurements under stated conditions \cite{r15}. Reference \cite{r16} propagates the uncertainty of a gated S-parameter through the inverse transform and the gate. That result gives the uncertainty against which a reconstructed gated response must be compared. Reference \cite{r17} maps received signal strength to the permittivity of liquids with a learned model. In that work every state is present in training. In this work each state is held out in turn. The target is the gated response. Section~7 reports the dependence of phase on magnitude that we measured on eleven configurations.

\section{Measurement Setup and Phantom Construction}

\begin{figure}[tbp]
\centering
\includegraphics[width=0.8\textwidth]{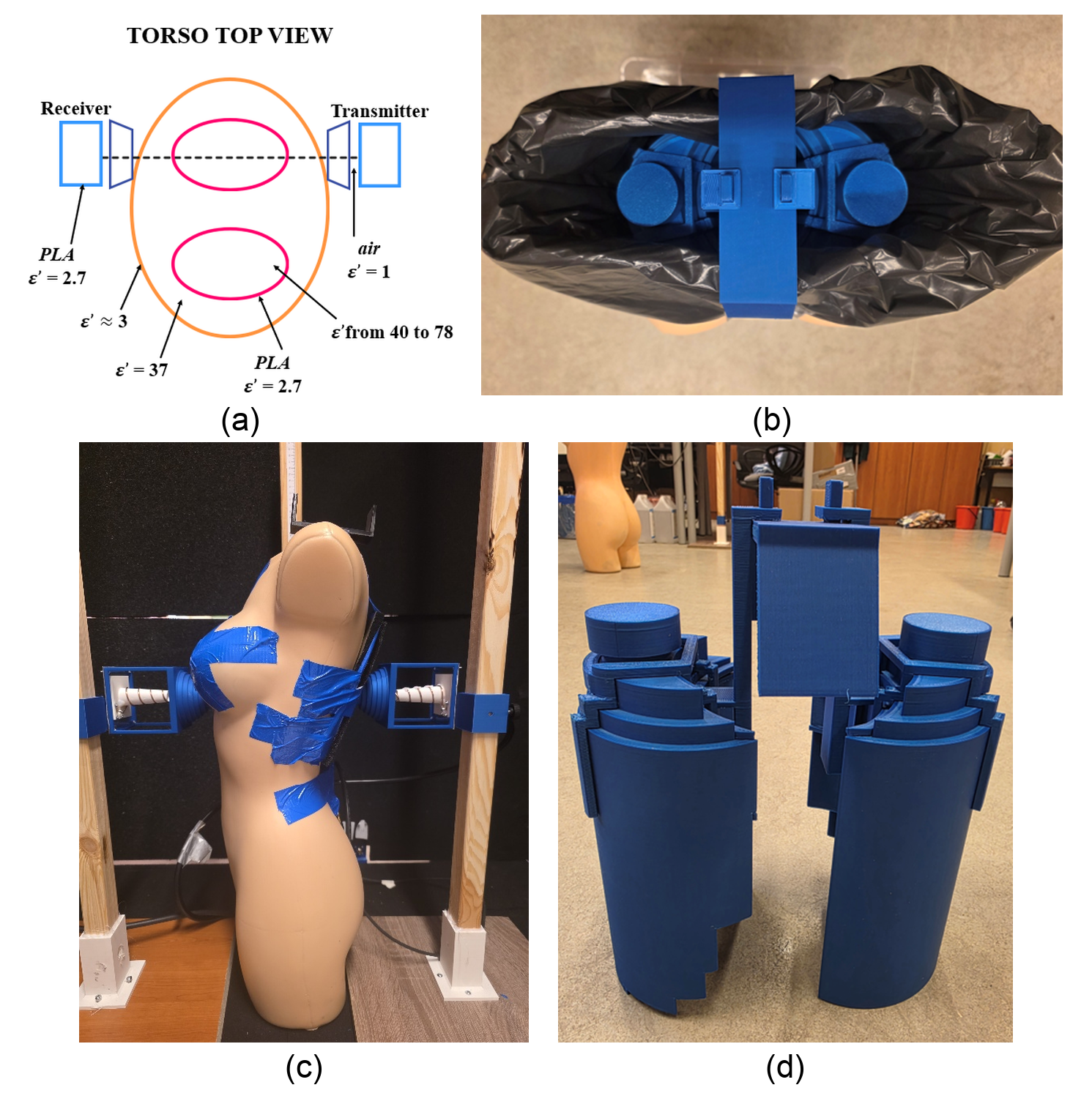}
\caption{Measurement phantom and setup. (a) Schematic of the phantom, top view. (b) Prepared measurement phantom, top view. (c) Measurement system, side view. (d) Lung-shaped compartments with positioning elements.}
\label{fig:1}
\end{figure}

The measurement setup followed the opposed two-antenna geometry used in an earlier study \cite{r9}, adapted here to the 2.0-3.0~GHz frequency range. Two helical antennas were positioned on opposite sides of the torso phantom and mounted on a rigid supporting frame (Fig.~\ref{fig:1}(c)). The frame maintained an antenna-to-phantom distance of 6~cm and constrained the relative position of the antennas and phantom throughout the experiment. The same mechanical arrangement was reproduced between measurement sessions, although small positioning differences could not be eliminated. Maintaining a well-defined geometry was important because the measured time-domain response, including the delays of the relevant propagation paths, depends on the spatial arrangement of the antennas and phantom.

S-parameters were acquired with a MegiQ VNA over 2.0-3.0~GHz at 1002 points, a step of about 1~MHz. The source power was set to $-10$~dBm and the sweep time was 1 s. A full two-port calibration was performed at the instrument ports, so the cables remain inside the measured path, and their delay appears in the time-domain response. However, the cables were rigid and treated in such a way that they were in a fixed position for every measurement.

The phantom consisted of an anatomically shaped tailor's mannequin shell filled with a homogeneous tissue-mimicking liquid with a real relative permittivity of approximately $\varepsilon'$ = 37 (Fig.~\ref{fig:1}). The homogeneous background was used to represent the chest wall, mediastinum and surrounding soft tissues by a single effective dielectric medium rather than by multiple discrete tissue layers. Although this constitutes a simplification of the dielectric structure of the human thorax, it provides a controlled background against which changes within the lung region can be studied independently. Importantly, unlike a fully homogeneous torso phantom, the present model included two anatomically positioned lung-shaped compartments manufactured from PLA. These compartments were filled with liquids of different dielectric properties, allowing the dielectric state of the lung-equivalent region to be varied while the surrounding torso-equivalent medium remained nominally unchanged. The phantom therefore represents a compromise between anatomical realism and experimental control, preserving the spatial distinction between the lungs and surrounding tissues while avoiding the additional variability introduced by a multilayer thoracic model.

This construction represents a deliberate compromise between anatomical realism and experimental control. The homogeneous background medium does not reproduce the complete multilayer dielectric structure of the human thorax. However, unlike a fully homogeneous torso phantom, the present model contains spatially distinct lung compartments whose dielectric properties can be modified independently. The experiment therefore preserves the principal dielectric contrast relevant to the present study while allowing controlled and repeatable modification of the lung-equivalent region.

The dielectric state of the lung compartments was controlled using water-ethanol mixtures of different compositions. The corresponding real part of the relative permittivity was measured at 2~GHz with a coaxial probe \cite{r18} and used to define the successive phantom states. Because the dielectric loss of water-ethanol mixtures depends on composition, frequency, and temperature, and was not independently measured during the experiment, it was not assigned a nominal value. The reported permittivity values should therefore be interpreted as nominal descriptors of the controlled dielectric states rather than as a complete characterization of the complex permittivity of each mixture. The measurements could be performed only after prior preparation of the phantom and the measurement setup. The torso was sealed to be watertight, providing unchanged conditions. The lungs were inserted inside. The next step involved proper positioning of the phantom and antennas inside the anechoic chamber, followed by the two-port calibration of the VNA at the instrument ports. After the calibration, the main chamber was filled with a homogeneous medium, and a subsequent measurement was performed during which the phantom's lungs were filled with air. Subsequently, the lungs were filled with mixtures representing various dielectric properties. Between subsequent measurements, the dielectric properties of the material inside the lungs were modified. Variety between signal propagation of each stage of filling lung-shaped components was measured.

The real relative permittivity of the liquid filling the lung compartments was varied from $\varepsilon'$ = 78.0 to $\varepsilon'$ = 47.8 in eleven states, with intermediate values of 72.9, 68.4, 64.5, 61.0, 58.0, 55.4, 53.1, 51.1, and 49.3 (Table~\ref{tab:1}). These states were generated sequentially while the dielectric properties of the surrounding torso-equivalent medium were kept nominally unchanged. The antenna arrangement and phantom position were maintained using the same mechanical setup and were reproduced as closely as possible between measurement sessions. Since the measurements were acquired over multiple days, small between-session variations in the exact antenna and phantom positioning could not be eliminated. The resulting dataset therefore includes limited realistic inter-session variability in addition to the controlled changes in the dielectric state of the lung-equivalent compartments.

\section{Data}

The primary dataset comprised 286 VNA sweeps acquired for eleven controlled dielectric states of the lung-equivalent compartments, with approximately 26 repeated measurements per state. Full two-port VNA measurements were stored in Touchstone (.s2p) format. The analyses presented in this work use the $S_{11}$ and $S_{21}$ responses.

The corresponding magnitude-only dataset acquired with the spectrum analyzer comprised 256 reflection-channel sweeps and 283 transmission-channel sweeps over the same set of phantom states (Table~\ref{tab:1}). The reflection dataset did not include measurements for the nominal $\varepsilon'$ = 68.4 state. The individual SA and VNA sweeps were not acquired as paired measurements; consequently, correspondence between the two instruments was defined at the level of the phantom state rather than on a sweep-by-sweep basis.

For model training and evaluation, the target signals were obtained by time-domain gating of the complex VNA measurements, using fixed gate parameters for each S-parameter as described in Section~4.1. Measurements from ten additional experimental configurations were not included in the reconstruction dataset and were used exclusively for the magnitude-phase dependence analysis presented in Section~7.

\begin{table}[tbp]
\caption{Lung-compartment filling steps and nominal real permittivity at 2~GHz.}
\label{tab:1}
\centering
\begin{tabular}{ccc}
\toprule
Step & Exchanged (mL) & $\varepsilon'$ at 2~GHz \\
\midrule
0 & 0 & 78.0 \\
1 & 100 & 72.9 \\
2 & 200 & 68.4 \\
3 & 300 & 64.5 \\
4 & 400 & 61.0 \\
5 & 500 & 58.0 \\
6 & 600 & 55.4 \\
7 & 700 & 53.1 \\
8 & 800 & 51.1 \\
9 & 900 & 49.3 \\
10 & 1000 & 47.8 \\
\bottomrule
\end{tabular}
\end{table}

\section{Method}

\subsection{Time-Domain Target Definition and Gate Placement}

The reconstruction target was defined as the magnitude spectrum obtained after time-domain gating of the complex VNA measurement. For each measured S-parameter, the complex frequency-domain spectrum acquired over 2.0-3.0~GHz at 1002 frequency points was transformed into the delay domain using an inverse fast Fourier transform (IFFT). No zero-padding or additional spectral window was applied before the transformation. The frequency spacing resulted in a delay-domain sampling interval of approximately 1~ns, while the effective temporal resolution was determined primarily by the 1~GHz measurement bandwidth.

Gate placement was guided by the expected propagation paths in the measurement geometry. The largest component of the delay-domain response was not assumed to correspond to the signal carrying the greatest information about the dielectric state of the lung-equivalent compartments. In reflection, the strongest early contributions are associated predominantly with the antenna response and the outer phantom boundary, whereas components involving propagation deeper within the phantom are expected at later delays. Consistently, the dominant $S_{11}$ response at approximately 24~ns showed only weak variation between phantom states, whereas a later component around 27~ns exhibited substantially stronger state-dependent variation.

A similar principle was applied to $S_{21}$. The strongest transmission component can contain contributions from shorter propagation paths, including propagation around the phantom rather than through its interior. Consequently, the transmission signal of interest was sought at a later delay, where the contribution of propagation through the phantom was expected to be greater. The known geometry and the observed delay-domain response were therefore used to define physically plausible search regions separately for $S_{11}$ and $S_{21}$.

Within these restricted delay regions, the exact gate center was selected using the Calinski-Harabasz (CH) statistic \cite{r19} calculated from the gated responses labeled by phantom dielectric state. The CH statistic quantifies the ratio of between-state to within-state dispersion and was used here to identify the delay at which the controlled dielectric states were most clearly separated. Importantly, the optimization was not performed over the complete delay-domain response, but only within the regions identified on the basis of the expected propagation paths.

The gate-selection procedure was performed once for each S-parameter using the measurements from the primary experimental configuration. Because the relevant propagation delays are determined predominantly by the antenna-phantom geometry, a single gate was subsequently fixed and applied to all dielectric states and all repeated sweeps within that configuration. The selected gate was centered at 27~ns with a span of 1.5~ns for $S_{11}$ and at 20~ns with a span of 2.0~ns for $S_{21}$.

A Kaiser window with $\beta = 6$ was used for gating. After selection of the desired delay-domain contribution, the gated signal was transformed back into the frequency domain. The magnitude of this gated complex spectrum constituted the target signal used for subsequent regression.

\subsection{Per-Device Reference Correction}

The VNA and SA measurements are not directly comparable on an absolute magnitude scale. The VNA provides calibrated complex S-parameters, whereas the SA records a magnitude-only response generated using its tracking generator. For reflection measurements, the SA signal is acquired through a VSWR bridge, while for transmission measurements it is measured through the direct tracking-generator-to-receiver path. Consequently, the measured magnitude includes frequency-dependent contributions from the respective instrument paths, including source level, cable insertion loss, and the response of the additional RF components.

Each device is therefore corrected once against a reference state. This is a correction in the sense of the International Vocabulary of Metrology (VIM) \cite{r20}. The only calibration in this work is the two-port VNA calibration of Section~2. Let $\hat{x}_d(f)$ be device $d$'s mean magnitude at the reference state $v^{*}$ and $\hat{y}_{\mathrm{VNA}}(f)$ the VNA mean gated magnitude at $v^{*}$. Training uses:

\begin{equation}
\tilde{x}_d^{(i)}(f) = x_d^{(i)}(f) - \hat{x}_d(f),
\label{eq:1}
\end{equation}

\begin{equation}
\tilde{y}^{(i)}(f) = y^{(i)}(f) - \hat{y}_{\mathrm{VNA}}(f),
\label{eq:2}
\end{equation}

and $\hat{y}_{\mathrm{VNA}}(f)$ is added back at inference.

The network therefore predicts changes relative to the reference state rather than the absolute gated magnitude spectrum. At inference, the reference gated VNA spectrum $\hat{y}_{\mathrm{VNA}}(f)$ is added back to the predicted relative response to recover the reconstructed gated spectrum on the VNA reference scale.

The reference correction is performed independently at every frequency bin and requires only a single phantom state common to both instruments. Individual SA and VNA measurements did not need to be paired on a sweep-by-sweep basis. Instead, measurements from both instruments were matched according to the corresponding phantom state. In the present dataset, the state with $\varepsilon'$ = 47.8 (1000~mL exchanged) was used as the reference state.

The correction is intended to remove the approximately state-independent, frequency-dependent difference between the measurement chains while preserving spectral changes associated with the dielectric state of the phantom. This assumes that the device-specific transfer characteristics remain approximately unchanged across the investigated phantom states.

The effect of the reference-based alignment on reconstruction performance is quantified separately in Section~6.

\subsection{Model, Loss, and Training}

The regressor is a fully-connected network, layer widths 1002, 256, 256, 1002, GELU activations, dropout 0.2, about $5.9 \times 10^{5}$ weights. Input is standardized per frequency bin, output by a single scalar over all bins. The gate couples every input bin to every output bin through the inverse Fourier kernel, so a fully connected network is appropriate. Deeper or wider variants brought no improvement. Optimization is AdamW \cite{r21} at $10^{-3}$ with weight decay $10^{-4}$, batch size 128, cosine schedule down to $10^{-5}$, early stopping with patience 10 on a held-out tenth of the training fold, at most 60 epochs.

Per-bin mean-squared error alone is not sufficient on SA input. A model can match the average per-state shape while the per-state curves remain incorrectly ordered; the eleven per-state means carry the ordering and a per-bin error does not constrain them. The training loss is therefore

\begin{equation}
L = (1-\alpha)\, L_{\mathrm{MSE}} + \alpha\, L_{\mathrm{aux}},
\label{eq:3}
\end{equation}

where $\alpha$ is annealed linearly from 0 to 0.3 over training so the auxiliary term in \eqref{eq:3} takes effect once predictions are roughly correct in scale. Within a batch, let $V$ be the states with at least two samples and $p_v(f)$, $g_v(f)$ the batch per-state mean prediction and target. Then

\begin{align}
L_{\mathrm{aux}} = {} & \Bigl\langle \sum_{a<b} \bigl[(p_a - p_b) - (g_a - g_b)\bigr]^2 \Bigr\rangle_f \nonumber \\
& + \Bigl\langle \sum_{a<b} \operatorname{relu}\bigl(-(p_a - p_b)\cdot\operatorname{sign}(g_a - g_b)\bigr) \Bigr\rangle_f .
\label{eq:4}
\end{align}

The first term of \eqref{eq:4} matches predicted inter-state differences to true ones at each frequency. The second is a pairwise hinge, nonzero exactly when a predicted pairwise difference has the wrong sign; it serves as a differentiable surrogate for the Spearman metric of Section~5.

Two training regimes are compared. The first trains on VNA sweeps only, at states other than the held-out one; SA data appear only at test time. The second adds SA sweeps at the same states. An SA sweep's target is the per-state mean of the VNA-gated curves at its state, since sweep indices do not pair across instruments. Training on both instruments forces SA and VNA inputs at a given state to the same output, which compensates the residual instrument transfer left after the reference subtraction. In both regimes the held-out state is absent from training.

The finite time gate imposes a characteristic spectral correlation scale of order $1/T$, where $T$ is the gate width; for a 2~ns gate this corresponds to approximately 500~MHz. The predicted spectra occasionally contained fluctuations over only a few frequency bins, substantially shorter than this scale. For visualization only, the predicted curves were therefore smoothed with a Gaussian kernel of $\sigma = 8$ bins. All quantitative metrics were calculated from the unsmoothed predictions, and the smoothing does not change any reported result.

\section{Evaluation}

Evaluation holds out one dielectric state at a time. For each state $v$ a fresh model is trained without any sweep at $v$ and predicts $v$; the eleven folds are combined into per-state means and overall metrics. A random sample-level split would give each test sweep about 25 training neighbors at the same state, and a stratified 80/20 split indeed returns Spearman 1.00 on $S_{21}$. Deployment presents dielectric states the model has not seen, and holding out each state in turn tests exactly this case. The reference spectra of Section~4.2 are treated as installation reference data and remain available in every fold, including the fold in which the reference state itself is held out; that fold withholds the training sweeps at the reference state while the correction stays in place. The $\varepsilon'$ = 47.8 reference correction necessarily enters the fold that holds out that level, and that fold has the largest shape error on both S-parameters.

Two metrics are reported. First, shape error is per-sweep mean-squared error in~dB$^2$ after removal of the constant offset. The offset is a single number per sweep, the mean over frequency of the difference between predicted and target gated magnitude in~dB. It corresponds to an overall level difference between the reconstruction and the VNA scale, of the kind that the installation correction removes. It is removed here because the present metric is intended specifically to quantify the frequency-dependent reconstruction error independently of a residual broadband level offset. It is subtracted from the error before squaring, so that shape error measures only the frequency-dependent part of the error. Second, preservation of the dielectric-state ordering was evaluated using the Spearman rank correlation between target and predicted per-state means, computed at each frequency bin. Per-bin values are combined as a median weighted by the target's across-state variance; unweighted, bins with no state-dependent signal would dominate. Spearman is used rather than Pearson because magnitude is monotone in the dielectric state but saturates at the low-permittivity end of the range. The Calinski-Harabasz statistic of predictions labeled by dielectric state is reported as a consistency check and is not optimized.

The repeatability of the VNA measurement sets the lower bound of the target error. Within a state, repeated sweeps have a standard deviation of 0.025~dB on $S_{11}$ and 0.45~dB on $S_{21}$ before gating. After gating the values are 0.014~dB and 0.027~dB (medians over frequency). The repeatability component is the shape error from gating one sweep with the phase of another repeat of the same state: $3 \times 10^{-4}$~dB$^2$ (Section~6.1). Substitution of the per-state mean phase gives 0.30~dB$^2$. This value bounds the spread between the sweeps of one state.

Each trained model was evaluated using two types of input. For VNA evaluation, a held-out VNA magnitude spectrum was used to reconstruct its corresponding gated VNA spectrum, providing a within-instrument reference for reconstruction performance. For SA evaluation, a held-out magnitude-only SA measurement was used to predict the mean gated VNA response of the corresponding phantom state. The latter represents the principal use case investigated in this work: reconstruction of a VNA-equivalent gated response from magnitude-only measurements.

\section{Results}

\subsection{Comparison With Explicit Phase Reconstruction}

The proposed reconstruction predicts the time-domain-gated magnitude directly from a magnitude-only input and therefore does not require phase at inference. An alternative strategy would be to first reconstruct the missing phase and subsequently apply conventional complex time-domain gating. We therefore examined whether phase could be recovered with an error small enough for this approach to be practical.

\begin{figure}[tbp]
\centering
\includegraphics[width=0.85\textwidth]{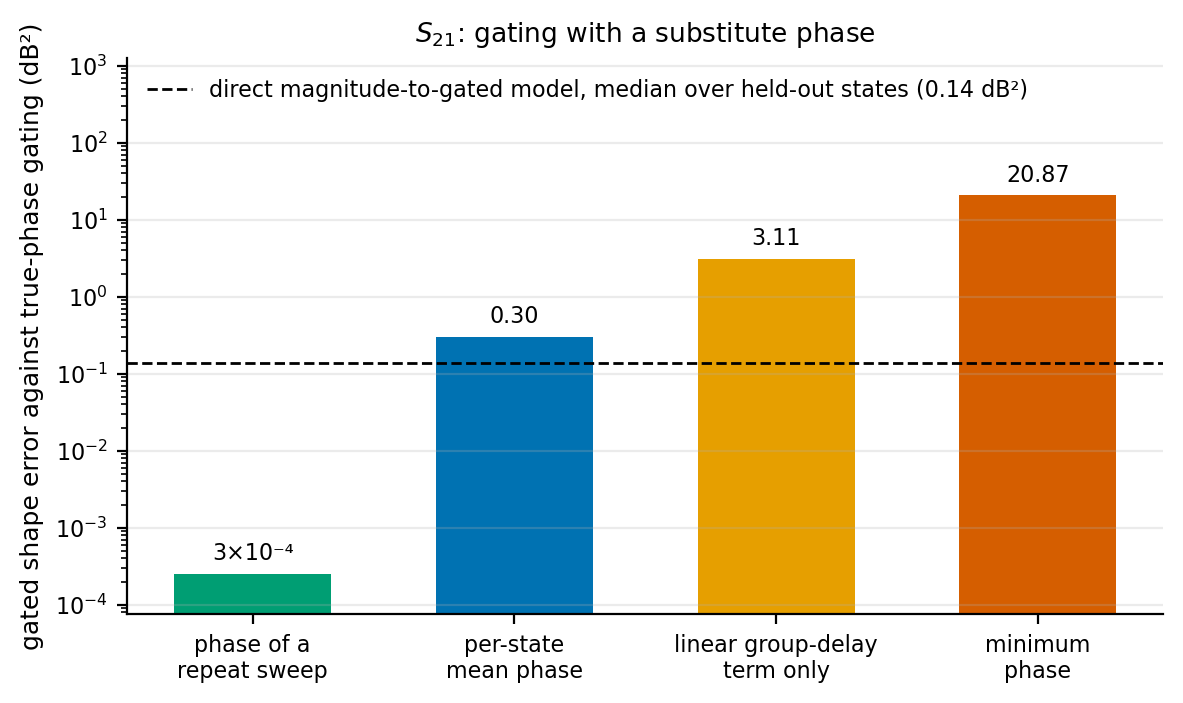}
\caption{$S_{21}$, gated shape error when the true magnitude is gated with a substitute phase, against gating with the true phase. The dashed line marks the direct magnitude-to-gated model at its within-VNA median over the held-out states.}
\label{fig:2}
\end{figure}

\begin{figure}[tbp]
\centering
\includegraphics[width=0.85\textwidth]{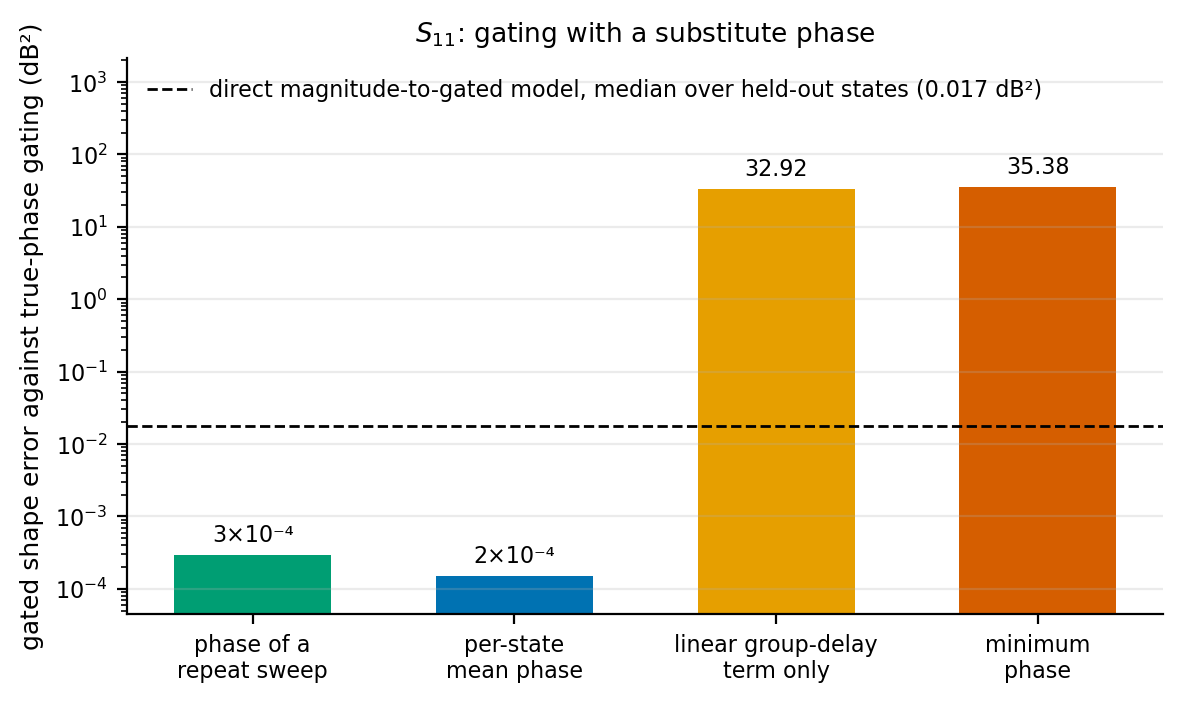}
\caption{$S_{11}$, gated shape error when the true magnitude is gated with a substitute phase, against gating with the true phase, as in Fig.~\ref{fig:2}. The dashed line marks the direct magnitude-to-gated model at its within-VNA median over the held-out states (0.017~dB$^2$).}
\label{fig:3}
\end{figure}

To establish how small the error of a reconstructed phase would have to be, we first added broadband phase noise of 0.05, 0.2, 0.5, and 1.0~rad RMS to the true phase and gated the result. The gated shape error on $S_{21}$ grew from $5 \times 10^{-3}$ through 0.1 and 0.5 to 4~dB$^2$, so a gated shape error below 0.1~dB$^2$ requires a phase error of at most about 0.2~rad RMS. We then asked whether measured phase meets this requirement, by gating the true magnitude with the phase of another repeat sweep of the same setup and dielectric state. This gives a gated shape error of $3 \times 10^{-4}$~dB$^2$ on $S_{21}$, and substituting the per-state mean phase gives 0.30~dB$^2$ (Figs.~\ref{fig:2} and \ref{fig:3}); the measured phase therefore carries enough information, and the remaining question is whether it can be predicted from magnitude. In the third step we trained a within-configuration predictor of the residual phase, the phase left after removal of the linear group-delay term, again with one state held out at a time. It reaches only about 1~rad RMS on $S_{21}$, and gating with its output gives a median shape error of 18~dB$^2$, against 0.12~dB$^2$ for the direct magnitude-to-gated model retrained in the same run. Finally, we retargeted the predictor at the per-state mean of the residual phase, an easier target. This reduces the difference, but even with that target phase known exactly the gated error is 0.29~dB$^2$, against 0.21~dB$^2$ for the direct model in that run, so a perfect predictor of the per-state mean phase would still be worse than the direct regression. The direct model is retrained as the baseline in each of these experiments, and its within-VNA median shape error over the held-out states on $S_{21}$ lies between 0.12 and 0.21~dB$^2$ across them (0.14~dB$^2$ in Fig.~\ref{fig:2}). The spread between runs is small next to the difference to any phase-based method. All results below use the direct regression.

Under the same protocol we also scored two methods from the SA magnitude to the gated target that use no learning. The first method estimates the phase as minimum phase (the Hilbert transform of the log magnitude) and then applies the gate. It gives a median shape error of 21.8~dB$^2$ on $S_{11}$ and 43.7~dB$^2$ on $S_{21}$. This is worse than a prediction of the reference curve for every state (0.44 and 2.87~dB$^2$). The measurement contains delayed arrivals and is therefore not minimum phase.

The second method is a per-bin linear map (ridge regression) from the reference-corrected SA magnitude to the gated target. It reaches 0.068~dB$^2$ on $S_{11}$ and 1.15~dB$^2$ on $S_{21}$. The network reaches 0.076 and 0.82~dB$^2$ (Table~\ref{tab:2}). With a fixed object the state-to-state change of the spectrum is close to linear in the magnitude. Without the smoothing step of Section~4 the linear map matches the network with VNA input on both S-parameters. The advantage of the network appears when the object changes. On the open breast data set of Section~6.4, with whole phantoms held out, its error is 30 percent below the linear map.

\subsection{Training on VNA Sweeps Only}

\begin{figure}[tbp]
\centering
\includegraphics[width=1.0\textwidth]{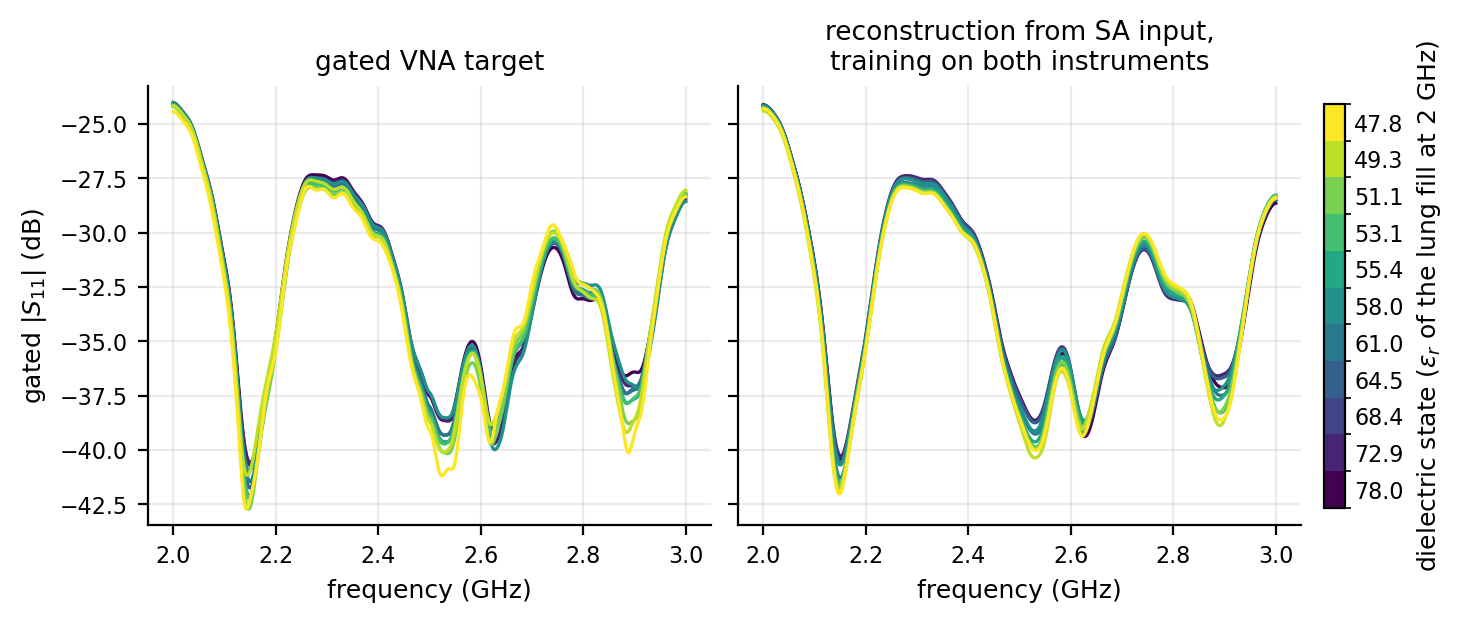}
\caption{$S_{11}$ per-state means. The left panel is the VNA-gated target; the right panel is the reconstruction from SA input with the full method (training on both instruments, ordering loss, smoothed).}
\label{fig:4}
\end{figure}

\begin{figure}[tbp]
\centering
\includegraphics[width=1.0\textwidth]{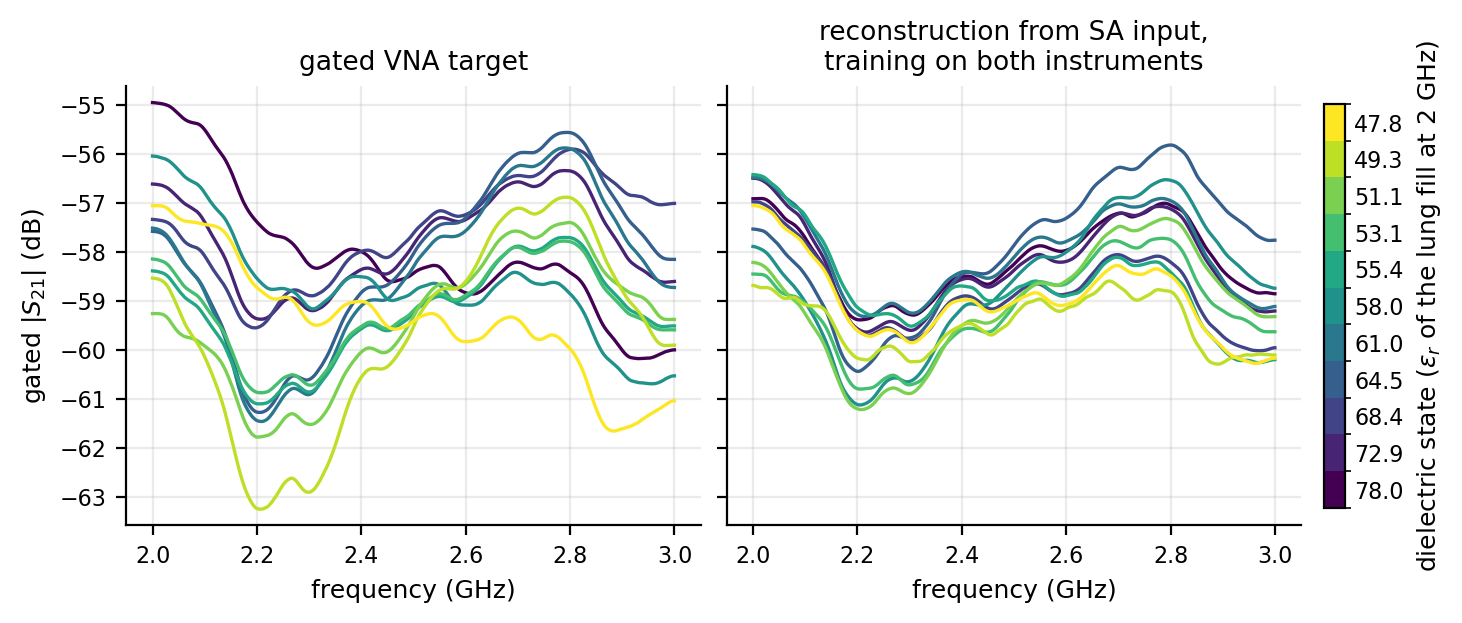}
\caption{$S_{21}$ per-state means, same display as Fig.~\ref{fig:4}. The three groups of dielectric states and their order are preserved across the change of instrument.}
\label{fig:5}
\end{figure}

We first evaluated whether a model trained exclusively on VNA magnitude spectra could be applied directly to measurements acquired with the SA. Within the VNA domain, reconstruction with one state held out at a time performed well. On $S_{11}$ the eleven per-state means of the gated target lie within a few tenths of a decibel of each other over most of the band and separate by two to three decibels only at the resonance dips near 2.15, 2.5, and 2.9~GHz (Fig.~\ref{fig:4}, left). Reconstruction from a held-out VNA sweep recovers both the common shape and the state-dependent spread at the dips, with a shape error of 0.037~dB$^2$ and an ordering of 0.90 (Table~\ref{tab:2}). On $S_{21}$ the target curves differ far more between states, by up to about six decibels near 2.2~GHz, and fall into three groups by permittivity (Fig.~\ref{fig:5}, left); the model trained on VNA sweeps reproduces the curves of a held-out state with a shape error of 0.052~dB$^2$ and an ordering of 0.86.

\begin{figure}[tbp]
\centering
\includegraphics[width=1.0\textwidth]{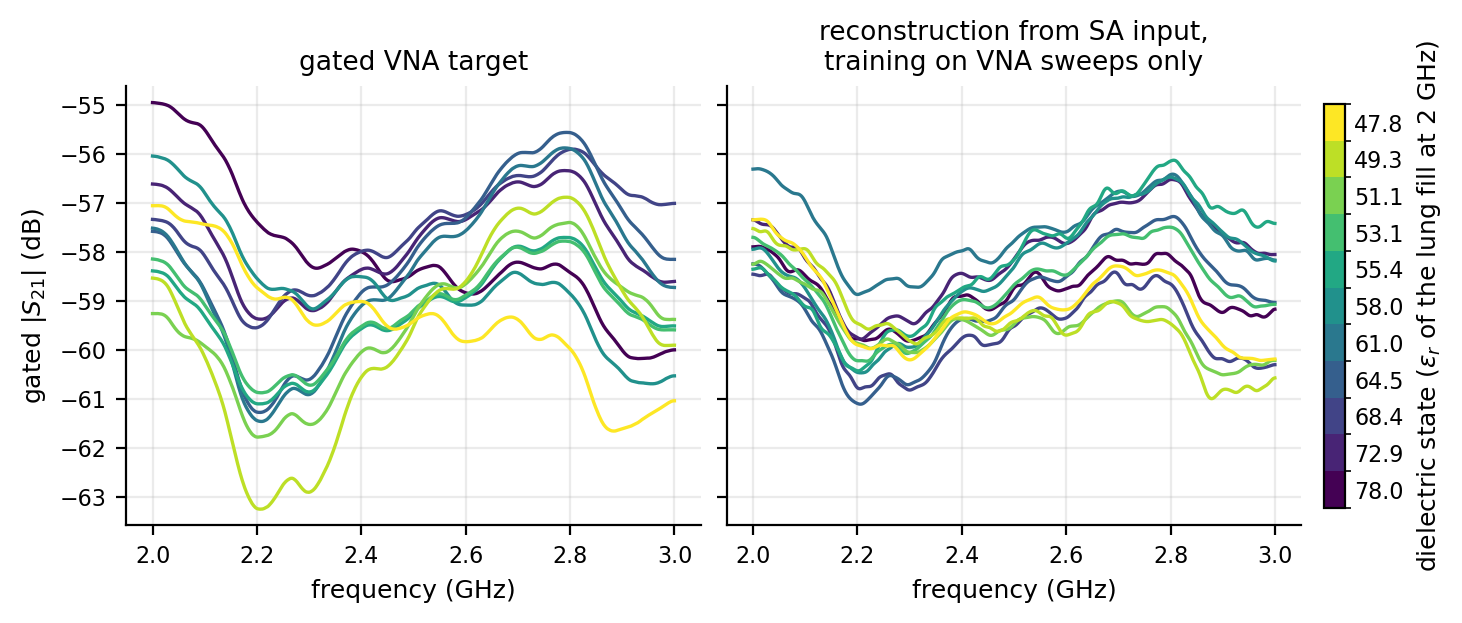}
\caption{$S_{21}$ per-state means, same display as Fig.~\ref{fig:5}; VNA-gated target left, reconstruction from SA input under training on VNA sweeps only right, ordering $-0.01$.}
\label{fig:6}
\end{figure}

\begin{figure}[tbp]
\centering
\includegraphics[width=1.0\textwidth]{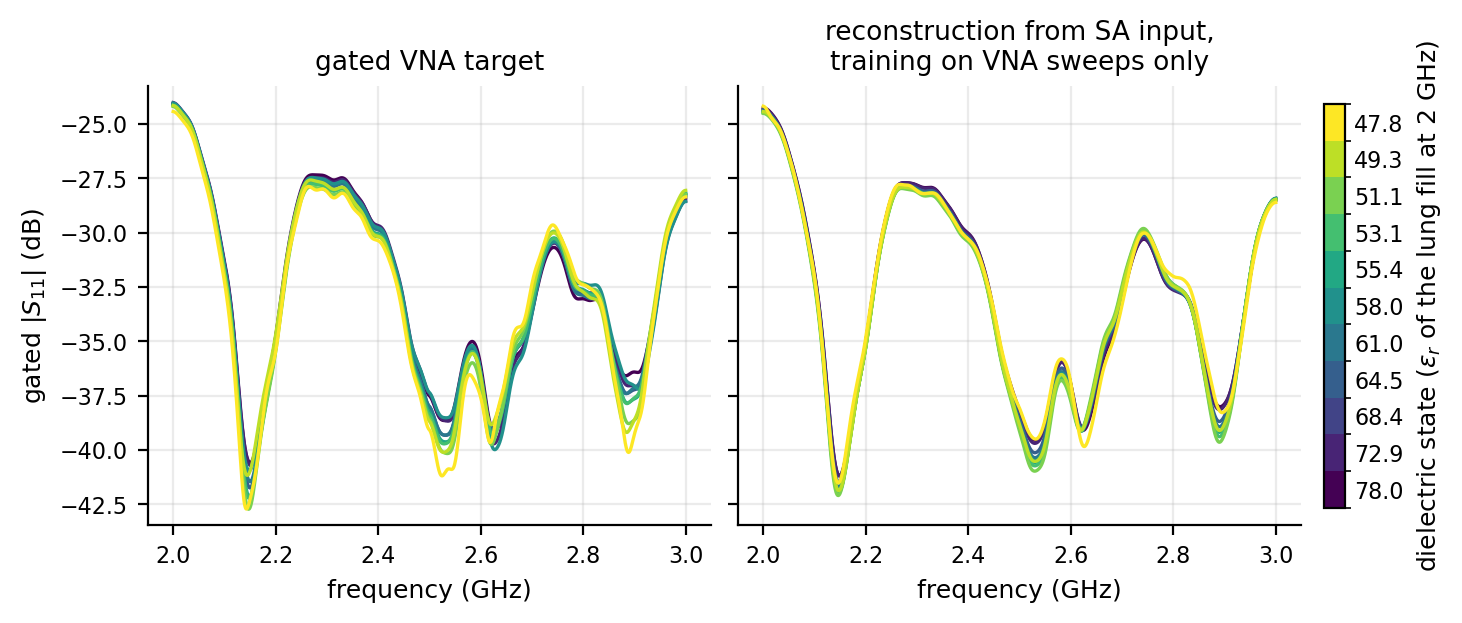}
\caption{$S_{11}$ per-state means, same display as Fig.~\ref{fig:4}; VNA-gated target left, reconstruction from SA input under training on VNA sweeps only right, ordering 0.59.}
\label{fig:7}
\end{figure}

The same models were then given SA sweeps as input. On $S_{21}$ the reconstruction fails: the ordering falls to $-0.01$, the curves of neighboring states cross each other over the whole band, and the three groups visible in the target are no longer recognizable (Fig.~\ref{fig:6}). On $S_{11}$ the common shape survives, because it is set by the antenna and barely depends on the state, but the state-dependent spread at the resonance dips is reproduced only in part and the ordering drops from 0.90 to 0.59 (Fig.~\ref{fig:7}). The shape error on SA input rises to 0.243~dB$^2$ on $S_{11}$ and 0.78~dB$^2$ on $S_{21}$ (Table~\ref{tab:2}), roughly an order of magnitude above the within-VNA values.

The SA spectrum contains the dielectric-state information. Trained from SA sweeps to per-state means on a stratified 80/20 split, all states seen, the model orders held-out sweeps at 1.00 for $S_{21}$ and 0.99 for $S_{11}$. Reconstruction fails when an unseen state is combined with a change of instrument.

\subsection{Training on Both Instruments}

Table~\ref{tab:2} compares the three training variants on both inputs and both S-parameters. Reconstructions with the full method are shown next to their targets in the right panels of Figs.~\ref{fig:4} and \ref{fig:5}.

\begin{table}[tbp]
\caption{Results with one dielectric state held out at a time. Shape error in~dB$^2$ and ordering as Spearman $\rho$, evaluated on VNA input and on SA input. The plain loss is per-bin mean-squared error alone.}
\label{tab:2}
\centering
\begin{tabular}{clcccc}
\toprule
S-param. & Training & VNA shape & VNA $\rho$ & SA shape & SA $\rho$ \\
\midrule
$S_{11}$ & VNA only & 0.037 & 0.90 & 0.243 & 0.59 \\
$S_{11}$ & both, plain loss & 0.047 & 0.87 & 0.076 & 0.82 \\
$S_{11}$ & both, ordering loss & 0.047 & 0.88 & 0.082 & 0.81 \\
$S_{21}$ & VNA only & 0.052 & 0.86 & 0.78 & $-0.01$ \\
$S_{21}$ & both, plain loss & 0.069 & 0.84 & 0.99 & 0.25 \\
$S_{21}$ & both, ordering loss & 0.061 & 0.86 & 0.82 & 0.28 \\
\bottomrule
\end{tabular}
\end{table}

For $S_{11}$, adding SA sweeps to training improves every metric on SA input. Shape error falls from 0.243 to 0.076~dB$^2$, ordering rises from 0.59 to 0.82, near the 0.88 on VNA input, and the degradation on VNA input (0.037 to 0.047~dB$^2$) is below run-to-run noise. The ordering loss and the plain loss are equivalent on $S_{11}$; the improvement comes from the added SA sweeps.

The behavior of $S_{21}$ is different. For $S_{21}$ the plain loss improves ordering and worsens shape. Ordering goes from $-0.01$ to 0.25 while shape error worsens from 0.78 to 0.99~dB$^2$, and the error on VNA input rises from 0.052 to 0.069~dB$^2$. The ordering loss restores shape to 0.82~dB$^2$ and brings ordering to 0.28 with no further degradation on VNA input.

The Spearman value understates what the $S_{21}$ reconstruction preserves. In Fig.~\ref{fig:5} the reconstructed curves on the right reproduce the clustering of the target on the left. The curves fall into three groups of dielectric states of high, intermediate, and low permittivity, and the groups sit in the correct order. Rank correlation counts every swap of adjacent permittivity steps inside a cluster as a miss, and the remaining error concentrates in such swaps, which is why ordering is 0.28 against 0.86 on VNA input. The Calinski-Harabasz statistic of predictions from SA input is about $8.0 \times 10^{3}$ for $S_{21}$ and $7.5 \times 10^{3}$ for $S_{11}$, two to four orders of magnitude above gating at the impulse maximum, so the cluster separation survives the change of instrument. Determining which group of dielectric states a measurement belongs to requires only this cluster-scale separation. Transmission-based detection deep in the torso relies on small $S_{21}$ changes \cite{r5}. For such use the reconstruction currently resolves the dielectric state at the cluster scale. Inference on the reconstructed curves is not evaluated in this work.

Input preprocessing did not change any conclusions. The training-set composition and the reference correction determine the result on SA input.

\subsection{External Validation on an Open Data Set}

The direct regression was also tested on data that we did not record. The University of Manitoba Breast Microwave Imaging Dataset (UM-BMID) \cite{r22} contains monostatic $S_{11}$ sweeps of breast phantoms in an air-filled chamber. Each scan has 72 antenna positions from 1 to 8~GHz, and each session has one sweep of the empty chamber. We used the second hardware generation: 1149 scans of 66 phantoms in 152 sessions, with 24 positions per scan.

The empty-chamber sweep of the session served as the complex reference. The input is the reference-corrected magnitude in~dB and the target is the gated magnitude of the reference-corrected complex sweep, as in Section~4. The Calinski-Harabasz criterion over the 66 phantoms placed the gate at 6.5~ns. Whole phantoms were held out in five folds, so every test sweep belongs to an object the model has not seen. Ordering was scored over the four breast-density classes of the set.

Table~\ref{tab:3} lists the median shape error of the held-out sweeps for two gate widths. The network of Section~4.3 reaches 1.8~dB$^2$ at a 0.5~ns gate. This is 30 percent below the per-bin linear map (2.5~dB$^2$). A third method reconstructs the complex sweep from the stored reference with a learned quadrature and then gates it. That method has the complex reference available and still scores 2.2~dB$^2$. The minimum-phase method fails (67~dB$^2$), as on our phantom.

A second random seed and the wider gate give the same ranking. The absolute errors are larger than on the thoracic phantom. The gate falls late in the response and separates the phantoms weakly, so the data set sets the level of the error. The ranking of the methods is the result that applies to our phantom. No open data set contains sweeps from a scalar instrument, so this test covers the VNA-magnitude input only.

\begin{table}[tbp]
\caption{Open breast data set with whole phantoms held out. Median shape error in~dB$^2$ of the held-out sweeps of UM-BMID generation two, five folds over 66 phantoms, for two gate widths. Ordering of the four density classes as Spearman $\rho$ at the 0.5~ns gate.}
\label{tab:3}
\centering
\begin{tabular}{lccc}
\toprule
Method & 0.5~ns gate & 1.0~ns gate & Ordering $\rho$ \\
\midrule
reference curve & 8.7 & 13.1 & $-0.8$ \\
per-bin linear map & 2.5 & 4.6 & 1.0 \\
network (this work) & 1.8 & 3.8 & 1.0 \\
minimum phase, then gate & 67 & 66 & $-0.8$ \\
stored reference, learned quadrature, then gate & 2.2 & 4.6 & 1.0 \\
\bottomrule
\end{tabular}
\end{table}

\section{Magnitude-Phase Dependence Within and Across Configurations}

The reconstruction method assumes that, within a given measurement configuration, the magnitude spectrum contains information related to the phase and therefore to the gated response. The results presented above suggest that this relationship is stronger for reflection than for transmission. We therefore examined the magnitude and phase relationship directly using VNA measurements, for which both magnitude and phase were available.

Two datasets were analyzed. The first consisted of 275 sweeps, which were acquired in the primary configuration used for the reconstruction experiments. The second combined ten further configurations, 6776 sweeps in total. These configurations differed in antenna pair, cabling, and placement, on the same 2.0-3.0~GHz grid. Magnitude was expressed in~dB. Phase was unwrapped and per-sweep linearly detrended to remove the group-delay slope. The analysis therefore focused on the remaining frequency-dependent phase variations rather than on the dominant delay introduced by the measurement path.

The analysis was performed in three ways: within the primary measurement configuration, after combining the ten additional configurations, and after combining the additional configurations while first subtracting the mean spectrum of each configuration. The last analysis reduces the influence of fixed differences between measurement setups and tests whether the magnitude-phase relationship remains in the spectral variations occurring within each configuration. Two statistics were computed in each setting. The distance correlation between the magnitude spectrum and the detrended phase \cite{r23} is zero only when the two are independent, so it tests whether a relation exists at all. The cross-validated $R^2$ of the phase regressed on 50 magnitude principal components (ridge regression, 5-fold \cite{r24}) measures how much of the phase a linear map from the magnitude components explains. A low value does not exclude a relation, since a nonlinear one can be present, so $R^2$ ought to be a lower bound on the dependence.

Within the primary configuration, the magnitude-phase relationship was clearly stronger for $S_{11}$ than for $S_{21}$. For $S_{11}$, the distance correlation was 1.00 and the cross-validated $R^2$ was 0.95. For $S_{21}$ the corresponding values were 0.75 and 0.64. Similar behavior was observed for the opposite ports, with reflection $S_{22}$ showing a stronger relationship than transmission $S_{12}$.

The same general trend was observed when measurements from multiple configurations were combined. The cross-validated $R^2$ was 0.97 for $S_{11}$ and 0.96 for $S_{22}$, compared with 0.52 for $S_{21}$ and 0.72 for $S_{12}$. The distance correlation shows the same ordering: 0.83 for $S_{11}$ and 0.89 for $S_{22}$ against 0.68 for $S_{21}$ and 0.77 for $S_{12}$. For reflection the value of $R^2$ close to 1 shows that phase is nearly a linear function of the magnitude components. For transmission the distance correlation of about 0.7 shows that a relation is present, while the $R^2$ of 0.5 to 0.7 shows that a linear map explains only part of it.

After subtraction of the mean spectrum of each configuration, the relationship remained present. For $S_{11}$, the cross-validated $R^2$ remained high at 0.86, whereas for $S_{21}$ it decreased to 0.47. This shows that the observed relationship is not caused only by fixed differences between measurement configurations. Variations in the spectral shape within a given configuration also contain information related to phase.

A possible physical explanation for the difference between reflection and transmission is the different propagation geometry. The $S_{11}$ response is mainly determined by the transmitting antenna and its local electromagnetic environment, including the nearby phantom region. In contrast, $S_{21}$ is formed by signals traveling between two spatially separated antennas and can contain contributions from several propagation paths. In the present setup, these may include both propagation through the phantom and propagation around its exterior.

Because these different contributions are combined coherently at the receiving antenna, small changes in antenna or phantom position can change their relative phase and therefore their constructive or destructive interference. The additional configurations used in this analysis differed in antenna pair, cabling, and positioning, so this type of geometrical variability may contribute to the weaker magnitude-phase relationship observed for $S_{21}$. This provides a plausible physical explanation, although the influence of positioning and individual propagation paths was not investigated independently in the present study.

The magnitude and phase analysis is therefore consistent with the reconstruction results. The stronger relationship observed for $S_{11}$ corresponds to the substantially better reconstruction obtained from magnitude-only SA measurements. In contrast, the more complex and geometry-sensitive propagation involved in $S_{21}$ is consistent with its larger reconstruction error and weaker preservation of dielectric-state ordering.

The measurements from the additional configurations were used only to characterize the magnitude-phase relationship. They do not demonstrate that the reconstruction model itself generalizes between different antenna, cable, or phantom configurations. Cross-configuration reconstruction remains outside the scope of the present study.

\begin{table}[tbp]
\caption{Magnitude-phase dependence. Distance correlation (dCor) and mean cross-validated $R^2$ of phase on magnitude, with detrended phase. Rows per S-parameter: the single configuration; ten configurations combined; ten configurations combined with each configuration's per-bin mean subtracted.}
\label{tab:4}
\centering
\begin{tabular}{clccc}
\toprule
S-param. & Regime & $n$ & dCor & CV $R^2$ \\
\midrule
$S_{11}$ & within (1 config.) & 275 & 1.00 & 0.95 \\
$S_{11}$ & across, combined & 6776 & 0.83 & 0.97 \\
$S_{11}$ & across, mean removed & 6776 & 0.56 & 0.86 \\
$S_{21}$ & within (1 config.) & 275 & 0.75 & 0.64 \\
$S_{21}$ & across, combined & 6776 & 0.68 & 0.52 \\
$S_{21}$ & across, mean removed & 6776 & 0.60 & 0.47 \\
$S_{12}$ & within (1 config.) & 275 & 0.74 & 0.37 \\
$S_{12}$ & across, combined & 6776 & 0.77 & 0.72 \\
$S_{12}$ & across, mean removed & 6776 & 0.76 & 0.68 \\
$S_{22}$ & within (1 config.) & 275 & 0.98 & 0.99 \\
$S_{22}$ & across, combined & 6776 & 0.89 & 0.96 \\
$S_{22}$ & across, mean removed & 6776 & 0.90 & 0.96 \\
\bottomrule
\end{tabular}
\end{table}

\section{Limitations}

The evaluation is on a single physical configuration. Generalization across configurations (different antennas, phantom mounting, cables) is a harder problem and is not addressed; training on both instruments has not been validated there. Both the gate selection and the training set use labeled data recorded at installation, so all claims are conditional on that reference data existing at install time.

The reference correction is required; without it, reconstruction from SA input fails (Section~4.2). Correction at several dielectric states, which could estimate the instrument transfer as a function of state instead of one offset, was not tested.

The experiments were performed on a controlled torso phantom and do not reproduce the full anatomical and dielectric complexity of the human thorax. The surrounding thoracic region was represented by a homogeneous dielectric medium, while the lungs were represented by PLA compartments filled with water-ethanol mixtures. The reported dielectric constant values describe nominal dielectric states at 2~GHz. The imaginary component of the permittivity was not independently measured, and possible effects of mixture temperature were not characterized. The investigated states should therefore be interpreted as controlled dielectric changes rather than as direct quantitative equivalents of specific physiological levels of pulmonary fluid accumulation.

Although the geometry was mechanically constrained, the data were collected over multiple measurement sessions and therefore include unavoidable small positioning and setup variations. The successful reconstruction under these conditions suggests that the method does not require perfectly identical sweep-to-sweep geometry. However, positioning robustness was not evaluated independently, and controlled perturbations of antenna or phantom position would be required to quantify this effect.

Finally, reconstruction of $S_{21}$ kept a larger shape error than that of $S_{11}$. The broader grouping of high-, intermediate-, and low-permittivity states was preserved, but ordering between neighboring states within these groups remained imperfect. The present work does not evaluate classification, estimation of dielectric state, or clinical inference from the reconstructed spectra. Whether the retained spectral information is sufficient for a diagnostic task therefore remains to be established.

\section{Conclusion}

With each dielectric state held out in turn, $S_{11}$ reaches shape error 0.047~dB$^2$ with ordering 0.88 from VNA input and 0.082~dB$^2$ with 0.81 from SA input; $S_{21}$ reaches 0.061~dB$^2$ with 0.86 from VNA input and 0.82~dB$^2$ with 0.28 from SA input. The method requires one calibrated configuration, an installation VNA grid for the gate and the reference correction, and one SA per deployed device. In deployment the steps follow the order used here: gate selection on the installation VNA grid, the per-device reference correction, and training on reference-subtracted VNA and SA sweeps with the annealed ordering loss. $S_{11}$ reconstruction reaches parity with VNA input; $S_{21}$ preserves the clustering of dielectric states with ordering inside clusters unresolved. On the open UM-BMID breast data set, with whole phantoms held out, the shape error of the network is 30 percent lower than that of a per-bin linear map. With a fixed object the linear map reaches the same error, so the nonlinear model is needed when the object changes. The present results are limited to phantom measurements and a single reconstruction configuration. Generalization across different measurement geometries and validation in more anatomically realistic or in vivo conditions remain subjects for future work.

\section*{Data availability}

The UM-BMID data set used in Section~6.4 is available at https://zenodo.org/records/5120981 \cite{r22}. The phantom sweeps and the code of this study are available from the corresponding author on request.

\section*{Funding}
This work was supported by Team NET [project number FENG.02.03-IP.05-0075/25]. The funder had no role in the study design, in the collection, analysis and interpretation of the data, in the writing of the report, or in the decision to submit the article for publication.

\section*{Acknowledgments}
This research was financed by Team NET under project number FENG.02.03-IP.05-0075/25. The authors gratefully acknowledge this support, which made it possible to develop and validate the system described in this work.

\section*{Declaration of competing interests}
The authors declare that they have no known competing financial interests or personal relationships that could have appeared to influence the work reported in this paper.

\end{document}